\documentclass[letterpaper,10pt,conference]{ieeeconf}  

\IEEEoverridecommandlockouts                              

\usepackage{graphicx,color}
\usepackage{epsfig} 
\usepackage{amsmath} 
\usepackage{amssymb}  
\usepackage{enumerate}
\usepackage{flushend}
\usepackage{cite}

\def\ddth{\mbox{$\ddot{\theta}$}}

\def\dth{\mbox{$\dot{\theta}$}}

\def\bh{\mbox{\boldmath$h$}}

\def\bq{\mbox{\boldmath$q$}}

\def\bJ{\mbox{\boldmath$J$}}

\def\bM{\mbox{\boldmath$M$}}

\def\bI{\mbox{\boldmath$I$}}

\def\blambda{\mbox{\boldmath$\lambda$}}
\def\bX{\mbox{\boldmath$X$}}

\def\bS{\mbox{\boldmath$S$}}

\def\bY{\mbox{\boldmath$Y$}}

\title{\LARGE \bf
Low-speed Limit Cycle Walking of Planar X-shaped Bipedal Robot with Special Properties
}

\author{
Fumihiko Asano and Cong Yan
\thanks{
Fumihiko Asano is with the Graduate School of Advanced Science and Technology, Japan Advanced Institute of Science and Technology, 1-1 Asahidai, Nomi, Ishikawa 923-1292, Japan {\tt\footnotesize fasano@jaist.ac.jp}%
}%
\thanks{
Cong Yan is with the Department of Mechanical Engineering, Ritsumeikan University, 1-1-1 Nojihigashi, Kusatsu Shiga 525-8577, Japan {\tt\footnotesize gansou@fc.ritsumei.ac.jp}%
}%
}

\begin{document}{

\maketitle
\thispagestyle{empty}
\pagestyle{empty}

\begin{abstract}
This paper proposes a novel planar semicircular-footed X-shaped bipedal robot in which the center of mass of each leg link is at the same position as the hip joint, and discusses a method for generating a stable limit cycle gait on level ground. First, we outline the assumptions of the robot model with four degrees of freedom, and develop the mathematical equations of motion and collision. Second, we set the relative hip-joint angle as a control output and design a control system to follow it to the desired trajectory specified as a fifth-order function of time. Third, we mathematically show that there are special properties: the angular velocity of the swing leg does not change before and after the collision, and the time-integral of the control input becomes zero in a steady gait. We also mathematically show that the model has the feature that approximate linearization is effective and the antigravity effect of semicircular feet can be utilized. Through numerical simulations, we verify the validity of the theoretical results, and analyze some fundamental gait properties such as convergence speed and mechanical energy restoration.

\end{abstract}
\setlength{\arraycolsep}{1.5pt}
\setlength{\abovedisplayskip}{6.1pt}
\setlength{\belowdisplayskip}{6.1pt}
\section{INTRODUCTION}

Compass-like biped robots \cite{Goswami1,Goswami2,Goswami3,Goswami4} have been studied for many years as the most basic model of passive bipedal walkers. Through the early studies, it was suggested that robotic bipedal locomotion is an essentially stable motion, and many natural and highly efficient bipedal robots have been realized by adding driving forces to the basic models. On the other hand, it has also been clarified that the effects of semicircular feet greatly contributes to the efficiency improvement of passive-dynamic and limit-cycle walking \cite{IROS2006,ICRA2007a,ICRA2007b}. Various high-efficiency legged locomotion utilizing them have been reported so far. Among them, extension to three-dimensional cases \cite{Tedrake,Collins,Science}, parametrically-excited dynamic bipedal walking \cite{Harata,Harata2,Harata3,IROS2007} are known as prominent ones.

In the field of robotics, compass-like biped robots are widely recognized as planar two-link robots with a total of three mass points at the hip joint, the midpoint of the stance (support) leg, and the midpoint of the swing (non-support) leg. Since the leg links extend only below the hip joint, the swing leg naturally swings forward due to the action of gravity during motion. By repeating the swinging motion while including collisions for stance-leg exchange, natural and efficient locomotion is generated without active control. There is the difficulty, however, that it is not easy to determine system parameters and initial conditions appropriately.

Based on the observation, in this paper we propose a novel planar X-shaped bipedal walker made by connecting two identical semicircular-footed leg links. This is nothing more than a simple two-link model with semicircular feet, but each link has a vertically symmetrical structure with the hip joint as the central point. Therefore, the above-mentioned natural swinging motion of the swing leg does not occur, and control of the swing-leg motion using the hip-joint torque is always required for stable gait generation. As one way to do this, we assume that the connecting joint is active and consider an output-following control system for achieving smooth movement of the relative angle between the two links for stable gait generation. We also discuss two special properties of the model; one is the antigravity effect of the semicircular feet, and the other is the property that the angular velocity of the swing leg does not change before and after the collision of stance-leg exchange. The former property results in a slow and leisurely locomotion, whereas the latter property results in the time integral of the control input for one steady step being zero. The validities of the proposed control method and theoretical results are investigated through numerical simulations.
\section{MODELING}

\subsection{Model Assumptions}

The planar X-shaped bipedal robot model with semicircular feet that walks on level ground is shown in Fig. \ref{fig2.01}. This model is made by connecting two identical semicircular-footed leg frames at their center points via a rotation axis, and has the following four degrees of freedom; $(x,z)$ is the end position of the stance leg, which is exactly equal to the center point of the semicircular foot; $\theta_1$ and $\theta_2$ are the angular positions of the stance leg and swing leg relative to the vertical or $Z$-axis. Let $L$ be the length from the rotation axis to the end position of each leg link, and $R$ be the radius of the semicircular foot attached to the end. Let the mass of each leg link be $m/2$, which is distributed by $m/4$ at two positions at a distance $r$ from the rotation axis. Therefore, the center of mass (COM) is at the same position as the rotation axis, and the inertia moment around it is $I=mr^2/2$. Hereinafter, the rotation axis is called the hip joint, and it is assumed that a rotational torque $u$ from the swing leg to the stance leg can be applied at the joint.

\begin{figure}[!t]
\centering
\vspace*{2mm}
\includegraphics[width=0.95\linewidth]{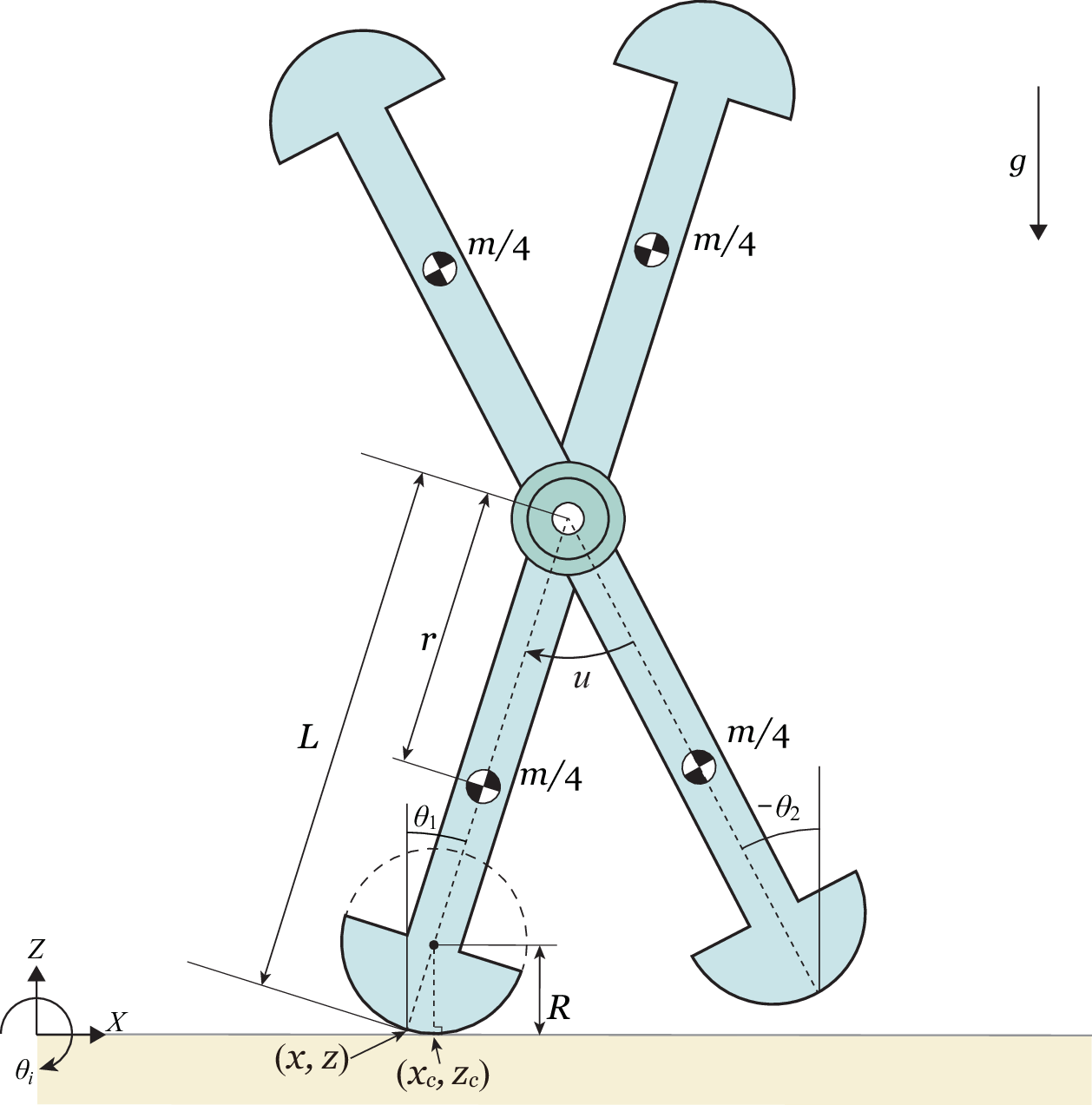}
\vspace*{-1mm}
\caption{Model of planar X-shaped bipedal robot made by connecting two identical semicircular-footed leg links}
\label{fig2.01}
\vspace*{-3mm}
\end{figure}

\subsection{Equation of Motion and Constraint Condition}

Let $\bq = \left[
\begin{array}{cccc}
x & z & \theta_1 & \theta_2
\end{array}
\right]^{\rm T}
$ be the generalized coordinate vector. The equation of motion for the X-shaped biped then becomes
\begin{equation}
\bM \ddot{\bq} + \bh = \bS u + \bJ_c^{\rm T} \blambda_c.
\label{eq2.01}
\end{equation}
The details of the terms on the left side are as follows.
\begin{eqnarray}
\bM &=& \left[
\begin{array}{cccc}
m & 0 & mL \cos \theta_1 & 0 \\
0 & m & -mL \sin \theta_1 & 0 \\
mL \cos \theta_1 & -mL \sin \theta_1 & mL^2 + I & 0 \\
0 & 0 & 0 & I
\end{array}
\right]\\
\bh &=& \left[
\begin{array}{c}
-mL \dth_1^2 \sin \theta_1 \\
-mL \dth_1^2 \cos \theta_1 + m g\\
-mgL \sin \theta_1 \\
0
\end{array}
\right], \ \ \bS = \left[
\begin{array}{c}
0 \\
0 \\
1 \\
-1
\end{array}
\right]
\label{eq2.02}
\end{eqnarray}
The velocity constraint conditions that the sole makes rolling contact with the floor surface without sliding, at the point $(x_c,z_c)$ in Fig. \ref{fig2.01}, are determined as follows.
\begin{eqnarray}
\tfrac{\rm d}{{\rm d}t}
\left(
x + R \sin \theta_1
\right) &=& R \dth_1 \\
\tfrac{\rm d}{{\rm d}t}
\left(
z + R \cos \theta_1
\right) &=& 0
\label{eq2.03}
\end{eqnarray}
By summarizing these, the constraint condition becomes
\begin{equation}
\bJ_c \dot{\bq} = {\bf 0}_{2 \times 1},
\label{eq2.04}
\end{equation}
where
\begin{equation}
\bJ_c = \left[
\begin{array}{cccc}
1 & 0 & R \left( \cos \theta_1 - 1 \right) & 0 \\
0 & 1 & - R \sin \theta_1 & 0
\end{array}
\right].
\label{eq2.05}
\end{equation}
This rolling contact constraint also allows the velocity vector to be expressed as a function only of angular velocity as
\begin{equation}
\dot{\bq} = 
\left[
\begin{array}{c}
\dot{x} \\
\dot{z} \\
\dth_1 \\
\dth_2
\end{array}
\right] = 
\left[
\begin{array}{cc}
R \left( 1 - \cos \theta_1 \right) & 0 \\
R \sin \theta_1 & 0 \\
1 & 0 \\
0 & 1
\end{array}
\right]
\left[
\begin{array}{c}
\dth_1 \\
\dth_2
\end{array}
\right].
\label{eq2.06}
\end{equation}
From Eq. (\ref{eq2.01}) and the time derivative of Eq. (\ref{eq2.04}), the undetermined multiplier vector $\blambda_c$, which means the ground reaction force vector, can be solved as
\begin{equation}
\blambda_c = -\bX_c^{-1} \left(
\bJ_c \bM^{-1} \left( \bS u - \bh \right) + \dot{\bJ}_c \dot{\bq}
\right),
\label{eq2.07}
\end{equation}
where $\bX_c := \bJ_c \bM^{-1} \bJ_c^{\rm T}$. By substituting Eq. (\ref{eq2.07}) into Eq. (\ref{eq2.01}) and arranging it, we obtain the simplified robot equation of motion as
\begin{equation}
\bM \ddot{\bq} = \bY_c \left( \bS u - \bh \right) - \bJ_c^{\rm T} \bX_c^{-1} \dot{\bJ}_c \dot{\bq},
\label{eq2.08}
\end{equation}
where $\bY_c := \bI_4 - \bJ_c^{\rm T} \bX_c^{-1} \bJ_c \bM^{-1}$.

\subsection{Collision Equation}

\begin{figure}[!b]
\centering
\vspace*{-3mm}
\includegraphics[width=0.95\linewidth]{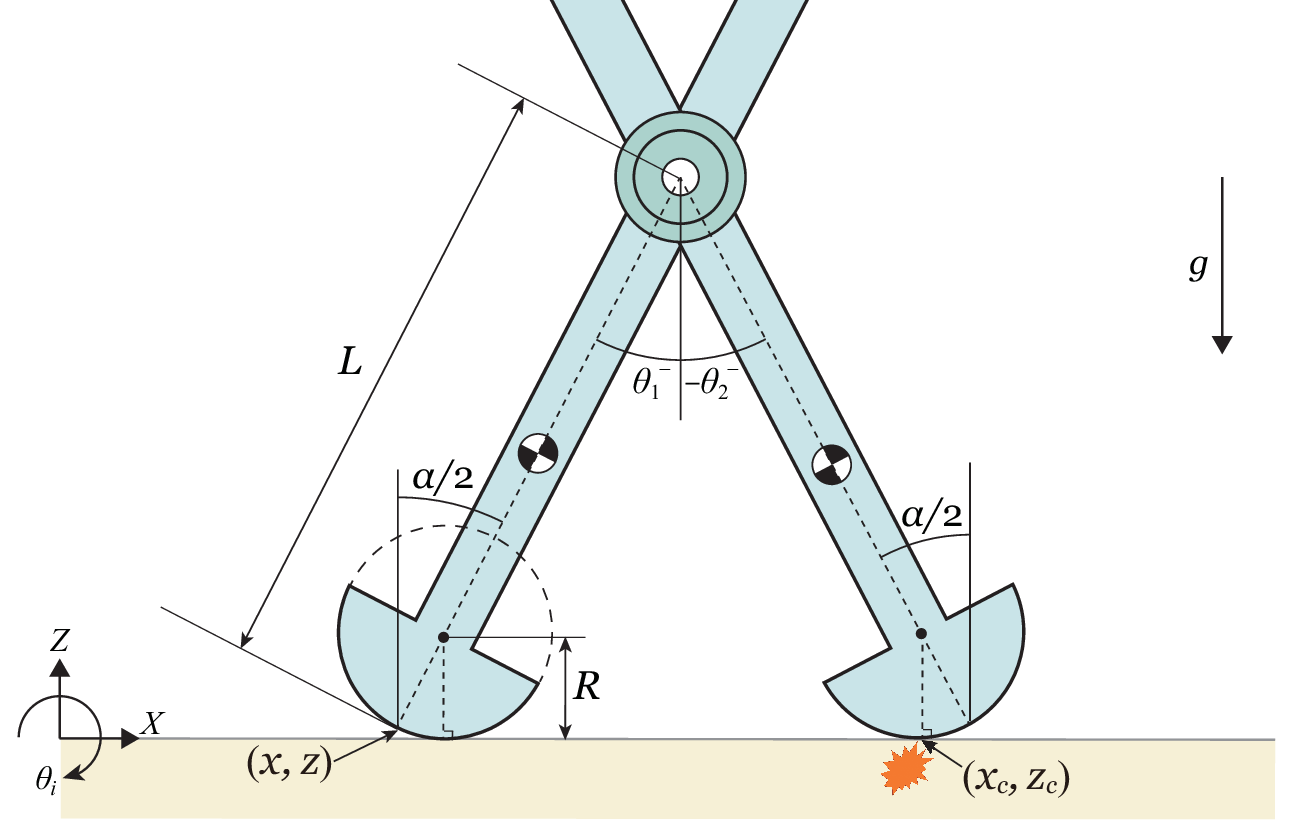}
\vspace*{-1mm}
\caption{Configuration at moment of collision}
\label{fig2.02}
\end{figure}

Fig. \ref{fig2.02} illustrates the configuration at the moment the fore foot lands. Note that, however, exchange of the coordinate system of the stance leg and swing leg is not considered. Assuming that the rolling contact constraint of the fore foot is established and the rear foot takes off immediately after the fore foot lands, the following completely inelastic collision equation is derived.
\begin{equation}
\bM \dot{\bq}^+ = \bM \dot{\bq}^- + \bJ_I^{\rm T} \blambda_I
\label{eq2.09}
\end{equation}
Here, the superscripts ``$-$'' and ``$+$'' shall denote immediately before and immediately after collision. The detail of $\bJ_I$ is described below. The following velocity constraint conditions should be satisfied immediately after collision at the point $(x_c,z_c)$ in Fig. \ref{fig2.02}.
\begin{eqnarray}
\tfrac{\rm d}{{\rm d}t}
\left(
x + L \sin \theta_1 - (L-R) \sin \theta_2
\right)^+ &=& R \dth_2^+ \\
\tfrac{\rm d}{{\rm d}t}
\left(
z + L \cos \theta_1 - (L-R) \cos \theta_2
\right)^+ &=& 0
\label{eq2.10}
\end{eqnarray}
By summarizing these conditions, we obtain
\begin{equation}
\bJ_I \dot{\bq}^+ = {\bf 0}_{2 \times 1},
\label{eq2.11}
\end{equation}
where
\begin{equation}
\bJ_I = \left[
\begin{array}{cccc}
1 & 0 & L \cos \theta_1^- & - (L-R) \cos \theta_2^- -R \\
0 & 1 & -L \sin \theta_1^- & (L-R) \sin \theta_2^-
\end{array}
\right],
\label{eq2.12}
\end{equation}
and $\theta_1^- = - \theta_2^- = \frac{\alpha}{2}$. By solving Eqs. (\ref{eq2.09}) and (\ref{eq2.11}) for the velocity vector immediately after impact, we obtain
\begin{equation}
\dot{\bq}^+ = \left(
\bI_4 - \bM^{-1} \bJ_I^{\rm T} \left(
\bJ_I \bM^{-1} \bJ_I^{\rm T}
\right)^{-1} \bJ_I
\right) \dot{\bq}^-.
\label{eq2.13}
\end{equation}
Note, however, that stance-leg exchange is not considered at this stage, and that the rear leg remains the stance leg. The velocity vector immediately before collision is specified as
\begin{equation}
\dot{\bq}^- = 
\left[
\begin{array}{c}
\dot{x}^- \\
\dot{z}^- \\
\dth_1^- \\
\dth_2^-
\end{array}
\right] = 
\left[
\begin{array}{cc}
R \left( 1 - \cos \frac{\alpha}{2} \right) & 0 \\
R \sin \frac{\alpha}{2} & 0 \\
1 & 0 \\
0 & 1
\end{array}
\right]
\left[
\begin{array}{c}
\dth_1^- \\
\dth_2^-
\end{array}
\right].
\label{eq2.14}
\end{equation}
Following Eqs. (\ref{eq2.13}) and (\ref{eq2.14}), the angular velocity immediately after impact is obtained as
\begin{eqnarray}
\hspace*{-5mm}
\dth_1^+ \! &=& \dth_1^-, 
\label{eq2.15a}
\\
\hspace*{-5mm}
\dth_2^+ \! &=& \!
\frac{
m F_1 \dth_1^- + I \dth_2^-
}{
m \left( L^2 - 2LR + 2R^2 \right) + 2m(L-R) R \cos \frac{\alpha}{2} + I
}, \\
\hspace*{-5mm}
F_1 \! &=&
R^2 + 2(L-R) R \cos \frac{\alpha}{2} + (L-R)^2 \cos \alpha.
\end{eqnarray}
By using these, the velocity vector immediately after impact that takes into account the leg exchange can be reset to
\begin{equation}
\dot{\bq}^+ = 
\left[
\begin{array}{cc}
R \left( 1 - \cos 
\left(
-\frac{\alpha}{2} 
\right)
\right) & 0 \\
R \sin 
\left(
-\frac{\alpha}{2}
\right) & 0 \\
1 & 0 \\
0 & 1
\end{array}
\right]
\left[
\begin{array}{cc}
0 & 1 \\
1 & 0
\end{array}
\right]
\left[
\begin{array}{c}
\dth_1^+ \\
\dth_2^+
\end{array}
\right].
\label{eq2.16}
\end{equation}
Note that $\dot{\bq}^+$ in Eq. (\ref{eq2.13}) and that in Eq. (\ref{eq2.16}) have the same notation, but differ in content. Eq. (\ref{eq2.15a}) describes that the angular velocity of the rear leg does not change at collision phase, and this is one of the special properties of the X-shaped biped.

After resetting the velocity vector, the position vector is also reset as
\begin{eqnarray}
\bq^+ &=& 
\left[
\begin{array}{c}
x^+ \\
z^+ \\
\theta_1^+ \\
\theta_2^+
\end{array}
\right] 
= \left[
\begin{array}{c}
x^- + L \sin \theta_1^- - L \sin \theta_2^- \\
z^- \\
\theta_2^- \\
\theta_1^-
\end{array}
\right] \nonumber \\
&=& \left[
\begin{array}{c}
x^- + 2 L \sin \frac{\alpha}{2} \\
R \left( 1 - \cos \frac{\alpha}{2} \right) \\
-\frac{\alpha}{2} \\
\frac{\alpha}{2}
\end{array}
\right],
\label{eq2.17}
\end{eqnarray}
and this completes the stance-leg exchange.
\section{CONTROL SYSTEM DESIGN AND SPECIAL PROPERTY OF ANGULAR VELOCITY}

\subsection{Basic Definitions}

In this paper, we always set the initial state of the robot to an upright stationary state, that is, $\bq (0) = {\bf 0}_{4 \times 1}$ and $\dot{\bq} (0) = {\bf 0}_{4 \times 1}$; this instant is defined as the 0-th collision. The next instant of stance-leg exchange or fore-foot landing is defined as the 1-st collision, and the motion between the 0-th and 1-st collisions is called the 0-th step motion. The subsequent collisions and step motions are contextually counted.

In the following, let $t$ be an absolute time variable with the starting time of walking set to zero, $i$ be the number of steps, and $\tau$ be a time variable which is reset to zero immediately after every collision. Let $T_i$ be the absolute time of the $i$-th collision, and $\Delta T_i := T_{i+1} - T_{i}$ be the $i$-th step period. If the generated gait is asymptotically stable, $\Delta T_i$ converges to a constant value, $\Delta T_{\infty}$.

Let $T_{\rm set}$ be the target settling-time for output following control of the relative hip joint. We assume that the inequality $T_{\rm set} \leq \Delta T_i$ always holds in every step as one of the conditions necessary to determine that a stable gait was generated; this is called the settling-time condition.

\subsection{Strict Output-following Control System Design}

We set the relative hip-joint angle, $\theta_H := \theta_1 - \theta_2 = \bS^{\rm T} \bq$, to the control output. The second-order derivative with respect to time of it becomes
\begin{eqnarray}
\hspace*{-5mm}
\ddth_H &=& \bS^{\rm T} \ddot{\bq} \nonumber \\
\hspace*{-5mm}
&=& \bS^{\rm T} \bM^{-1} \bY_c \left( \bS u - \bh \right) 
- \bS^{\rm T} \bM^{-1} \bJ_c^{\rm T} \bX_c^{-1} \dot{\bJ}_c \dot{\bq}.
\label{eq3.03}
\end{eqnarray}
From this, we can determine the control input that achieves $\ddth_H = v$ as $u = A^{-1} \left( v + B \right)$, where $A := \bS^{\rm T} \bM^{-1} \bY_c \bS$ and $B := \bS^{\rm T} \bM^{-1} 
\left( \bY_c \bh + \bJ_c^{\rm T} \bX_c^{-1} \dot{\bJ}_c \dot{\bq} \right)$. Among the desired time trajectories of $\theta_H$, we choose the following fifth-order time function as the simplest one that satisfies a total of six initial and terminal conditions.
\begin{equation}
\theta_{H{\rm d}}(\tau) = 
\left\{
\begin{array}{ll}
\sum_{k=0}^{5} a_k \tau^k & 
\left( 0 \leq \tau < T_{\rm set} \right)
\\
\alpha & 
\left( \tau \geq T_{\rm set} \right)
\end{array}
\right.
\label{eq3.06}
\end{equation}
For achieving strict output-following control that maintains the trajectory tracking error at zero, the angular position and angular velocity of the relative hip joint immediately after the $i$-th collision, $\theta_{Hi}^+$ and $\dth_{Hi}^+$, are reset to the target initial and terminal conditions each time as follows.
\begin{eqnarray}
\hspace*{-7mm}
& & 
\theta_{H{\rm d}}(0) = \theta_{Hi}^+, \ \ 
\dth_{H{\rm d}}(0) = \dth_{Hi}^+, \ \ 
\ddth_{H{\rm d}}(0) = 0 \\
\hspace*{-7mm}
& & 
\theta_{H{\rm d}}(T_{\rm set}) = \alpha, \ \ 
\dth_{H{\rm d}}(T_{\rm set}) = 0, \ \ 
\ddth_{H{\rm d}}(T_{\rm set}) = 0
\label{eq3.07}
\end{eqnarray}
The coefficients $a_k$ are then determined as
\begin{eqnarray}
a_5 &=& -\frac{3 \left(
\dth_{Hi}^+ T_{\rm set} + 2 \theta_{Hi}^+ - 2 \alpha
\right)}{T_{\rm set}^5}, \\
a_4 &=& \frac{8
\dth_{Hi}^+ T_{\rm set} + 15 \theta_{Hi}^+ - 15 \alpha
}{T_{\rm set}^4}, \\
a_3 &=& -\frac{2 \left(
3 \dth_{Hi}^+ T_{\rm set} + 5 \theta_{Hi}^+ - 5 \alpha
\right)}{T_{\rm set}^3}, \\
a_2 &=& 0, \ \ a_1 = \dth_{Hi}^+, \ \ a_0 = \theta_{Hi}^+.
\label{eq3.08}
\end{eqnarray}
The relative hip-joint angle is then strictly and smoothly controlled from $\theta_{Hi}$ to $\alpha$ during the $i$-th step motion, and after $\tau = T_{\rm set}$ the X-shaped biped falls down as a 1-DOF rigid body.

\subsection{Special Property of Angular Velocity}

According to the above strict output-following control, the following relationship holds immediately before the $i$-th collision.
\begin{equation}
\dth_1 \left( T_{i}^- \right) = \dth_2 \left( T_{i}^- \right)
\label{eq3.09}
\end{equation}
Considering that the rear leg that was the stance leg immediately before the collision switches to the swing leg immediately after the collision, the following relationship also holds.
\begin{equation}
\dth_1 \left( T_{i}^- \right) = \dth_2 \left( T_{i}^+ \right)
\label{eq3.10}
\end{equation}
Following, Eqs. (\ref{eq3.09}) and (\ref{eq3.10}), we obtain
\begin{equation}
\dth_2 \left( T_{i}^- \right) = \dth_2 \left( T_{i}^+ \right),
\label{eq3.11}
\end{equation}
and this is the special property of the X-shaped biped. The almost linear biped also has the same property \cite{Kiefer,Spong}.

Assuming that the generated gait completely converges to the limit cycle after the $i$-th step, the angular velocities of the swing leg immediately before and immediately after impact satisfy
\begin{equation}
\dth_2 \left( T_{i}^{\pm} \right) = \dth_2 \left( T_{i+1}^{\pm} \right) = \dth_2 \left( T_{i+2}^{\pm} \right) = \cdots
= \dth_2 \left( T_{\infty}^{\pm} \right).
\label{eq3.12}
\end{equation}
The swing-leg dynamics is specified as
\begin{equation}
I \ddth_2 = -u,
\label{eq3.13}
\end{equation}
and the time integral of it for the $i$-th step becomes
\begin{equation}
I 
\left(
\dth_2 \left( T_{i+1}^- \right) - \dth_2 \left( T_{i}^+ \right)
\right) = -\int_{T_i^+}^{T_{i+1}^-} u \ {\rm d}t.
\label{eq3.14}
\end{equation}
Considering Eq. (\ref{eq3.12}), the value of Eq. (\ref{eq3.14}) is zero. In the next section, we define
\begin{equation}
J_i := \int_{T_{i}^+}^{T_{i+1}^-} u \ {\rm d}t,
\label{eq3.15}
\end{equation}
and use it to evaluate the convergence of the generated gait.
\section{TYPICAL GAIT AND SPECIAL PROPERTY OF SEMICIRCULAR FOOT}


Fig. \ref{fig4.01} shows the simulation results of walking motion of the X-shaped biped on level ground where $T_{\rm set} = 1.0$ [s] and $\alpha = \frac{\pi}{4}$ [rad]. The physical parameters were also chosen as the values listed in Table \ref{table01}. We can see that the X-shaped biped started walking from an upright stationary state and generated a stable limit cycle gait according to the strict output-following control method. From Fig. \ref{fig4.01}(c), we can see that the relative hip-joint angle or control output is successfully controlled from $-\alpha$ to $\alpha$ during every single-limb support phase. From Fig. \ref{fig4.01}(d), we can see that the relative angular velocity of the hip joint becomes a slightly negative value, and then it is controlled to zero through acceleration and deceleration.

\begin{figure}[!t]
\vspace*{1mm}
\small
\centering
\scalebox{0.66}{
\input{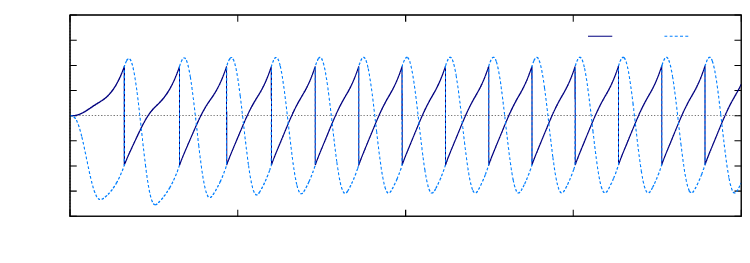}}
{\footnotesize (a) Angular position}
\scalebox{0.66}{
\input{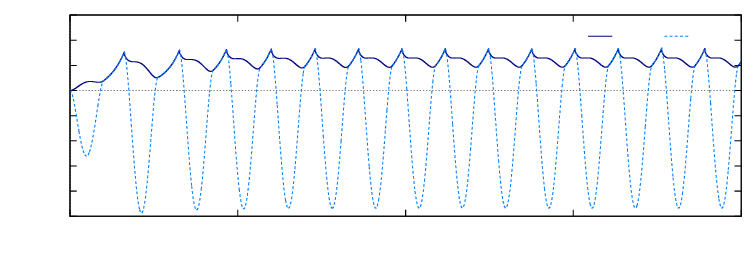}}
{\footnotesize (b) Angular velocity}
\scalebox{0.66}{
\input{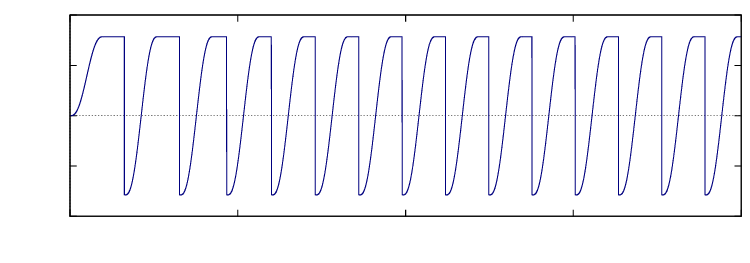}}
{\footnotesize (c) $\theta_H$}
\scalebox{0.66}{
\input{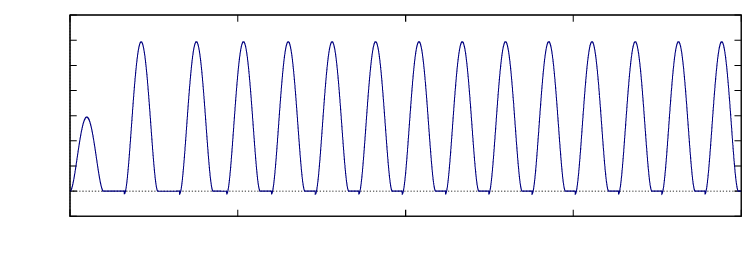}}
{\footnotesize (d) $\dth_H$}
\scalebox{0.66}{
\input{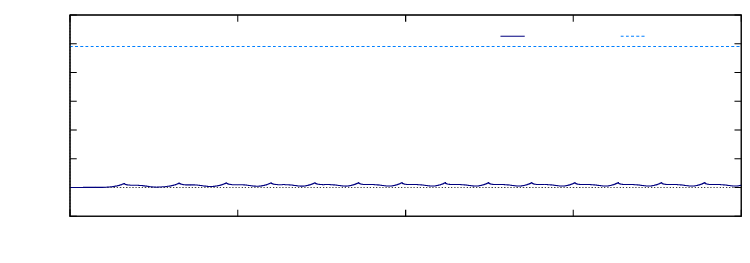}}
{\footnotesize (e) $R \dth_1^2$ and gravity}
\scalebox{0.66}{
\input{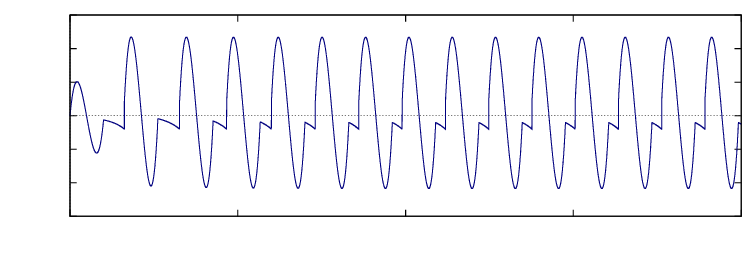}}
{\footnotesize (f) Control input}
\vspace*{-1mm}
\caption{Simulation results of walking motion of X-shaped biped on level ground}
\label{fig4.01}
\end{figure}

\begin{table}[!b]
\caption{Physical parameters of X-shaped biped}
\label{table01}
\centering
\renewcommand{\arraystretch}{1.3}
\begin{tabular}{lcc} \hline
$m$ & 2.0 & kg \\
$r$ & 0.5 & m \\
$I \ \left( = m r^2 / 2 \right)$ & 0.25 & kg$\cdot$m$^2$ \\
$L$ & 1.0 & m \\
$R$ & 0.5 & m \\
\hline
\end{tabular}
\end{table}

\begin{figure}[!b]
\vspace*{-2mm}
\small
\centering
\scalebox{0.66}{
\input{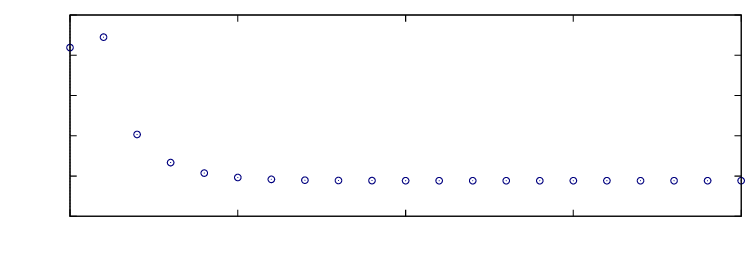}}
{\footnotesize (a) Step period}
\scalebox{0.66}{
\input{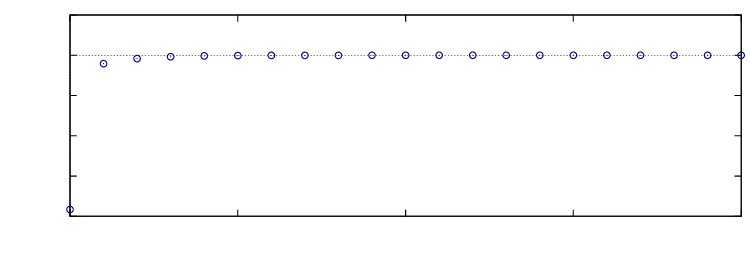}}
{\footnotesize (b) $J_i$}
\scalebox{0.66}{
\input{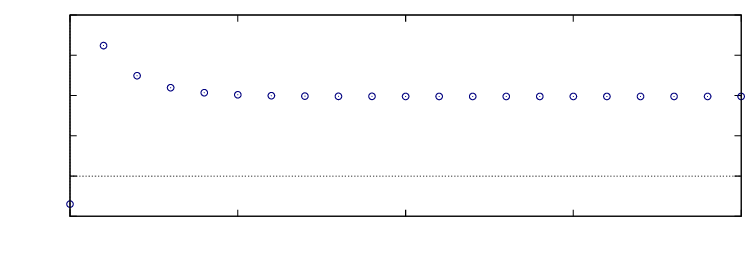}}
{\footnotesize (c) $\Delta E_i$}
\vspace*{-1mm}
\caption{Evolution of gait descriptors with respect to number of steps}
\label{fig4.02}
\end{figure}

Here we discuss another special property of the X-shaped biped with semicircular feet. The stance-leg dynamics is specified as
\begin{equation}
\ddth_1 = \frac{u + m(L-R) \left( g + R \dth_1^2 \right) \sin \theta_1}{m \left(
L^2 - 2LR + 2R^2
\right) + 2m(L-R)R \cos \theta_1 + I}.
\label{eq5.01}
\end{equation}
As seen in Fig. \ref{fig4.01}(e), the value of $R \dth_1^2$ is sufficiently smaller than that of $g = 9.81$ [m/s$^2$] during motion. Considering the approximations $g \gg R \dth_1^2$, $\sin \theta_1 \approx \theta_1$ and $\cos \theta_1 \approx 1$, Eq. (\ref{eq5.01}) can be arranged to
\begin{equation}
\ddth_1 \approx \frac{u + m \hat{g} L\theta_1}{m L^2 + I},
\label{eq5.02}
\end{equation}
where $\hat{g} := g \left( 1 - \frac{R}{L} \right)$. Eq. (\ref{eq5.02}) is equivalent to the linearized motion of a point-footed X-shaped biped under weak gravity. Therefore, the step period is slightly longer than usual and the generated walking motion is relaxed. It is for this reason that $T_{\rm set}$ was set to a slightly longer value of $1.0$ [s]. The use of semicircular feet has been proposed since McGeer's early studies \cite{McGeer0,McGeer,McGeer2}, but the antigravity effect mentioned above has never been pointed out before.

The above simplification of the stance-leg motion is possible only in models where the COM of each leg link is located at the same position as the hip joint, such as the model proposed in this paper, almost linear biped \cite{Kiefer,Spong}, and rimless-wheel-like walker formed by two cross-shaped frames \cite{ICRA2010}. Our X-shaped biped has a swing leg, but the dynamics is the same as an underactuated rimless wheel with an inertial disk.

As seen in Fig. \ref{fig4.01}(f), the control input, $u$, is positive in the first half of every single-limb support phase and is negative in the second half. According to the method, its time integral should converge to zero as the motion converges to the stable limit cycle. To confirm this, we plot the discrete metrics. Fig. \ref{fig4.02} shows the evolution of the gait descriptors; (a) is the step period, (b) the $J_i$, and (c) the restored mechanical energy for the $i$-th step, $\Delta E_i$. From the results, we can see that every descriptor converged to a constant value. The restored mechanical energy is determined as
\begin{equation}
\Delta E_i := \int_{T_i^+}^{T_{i+1}^-} \dth_{H} u \ {\rm d}t,
\label{eq5.03}
\end{equation}
and this also converged to a positive constant. The steady values of the step period and restored mechanical energy were $\Delta T_{\infty} = 1.28837$ [s] and $\Delta E_{\infty}= 0.0989367$ [J].

In a walking motion achieving constraint on impact posture, the following recurrence formula holds.
\begin{equation}
K_{i+1}^- = \varepsilon K_i^- + \Delta E_i
\label{eq5.04}
\end{equation}
Here, $K_i^-$ is the kinetic energy immediately before the $i$-th collision, and $\varepsilon := K_i^+ / K_i^-$ is the energy-loss coefficient. $\varepsilon$ can be analytically derived but is very complicated, so we omit the detail here. The value of $\varepsilon$ becomes constant if the constraint on impact posture is achieved. In addition, it monotonically increases as $R$ increases, and the minimum and maximum values are as follows.
\begin{eqnarray}
\lim_{R \rightarrow 0} \varepsilon &=& \frac{\left( m L^2 \cos \alpha + I \right)^2 + \left( m L^2 + I \right) I}{
\left( m L^2 + I \right) \left( m L^2 + 2I \right)
} \\
\lim_{R \rightarrow L} \varepsilon &=& 1
\end{eqnarray}
Fig. \ref{fig4.0x} plots $\varepsilon$ versus $R$ for eleven values of $r$. We can see that $\varepsilon$ monotonically increases with increasing $R$, and reaches 1 when $R = L$. $\varepsilon = 1$ means that kinetic energy loss at collision becomes zero, which in other words is perfect wheel rotational motion without collision.

\begin{figure}[!b]
\vspace*{-2mm}
\small
\centering
\scalebox{0.66}{
\input{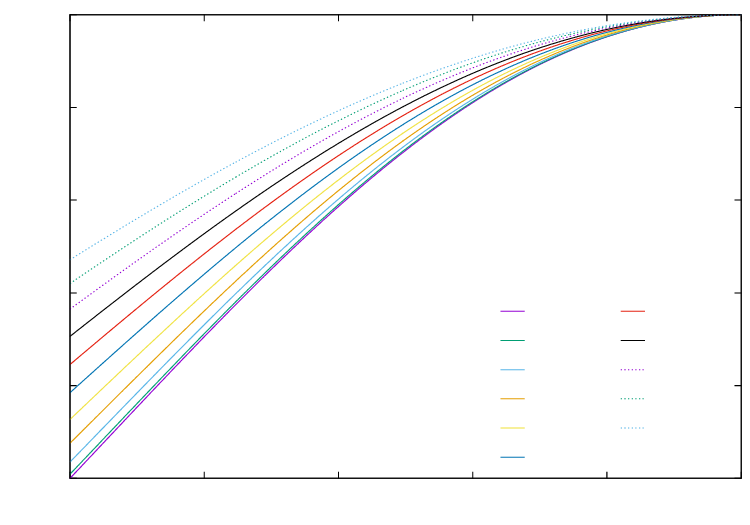}}
\vspace*{-1mm}
\caption{$\varepsilon$ versus $R$ for eleven values of $r$ where $\alpha = \frac{\pi}{4}$ [rad]}
\label{fig4.0x}
\end{figure}

If the generated walking motion is asymptotically stable, $K_i^-$ converges to
\begin{equation}
K_{\infty}^- = \lim_{i \rightarrow \infty} K_{i}^- = 
\frac{\Delta E_\infty}{1 - \varepsilon},
\label{eq5.05}
\end{equation}
where
\begin{equation}
\Delta E_\infty := \lim_{i \rightarrow \infty} \Delta E_i
\end{equation}
is the steady restored mechanical energy. It is easy to prove the asymptotic stability of the walking motion if $\Delta E_i$ is constant along with $\varepsilon$, but in this case it is difficult because the $\Delta E_i$ remains as a variable.

Fig. \ref{fig4.03} shows the phase-plane plot of the steady walking motion, and Fig. \ref{fig4.04} shows the stick diagram for one-step motion of it. Since the trajectory of the swing leg in Fig. \ref{fig4.03} jumps parallel to the horizontal axis at the moment of collision, it can be confirmed that the property of Eq. (\ref{eq3.12}) holds true.

\begin{figure}[!t]
\vspace*{1mm}
\small
\centering
\scalebox{0.66}{
\input{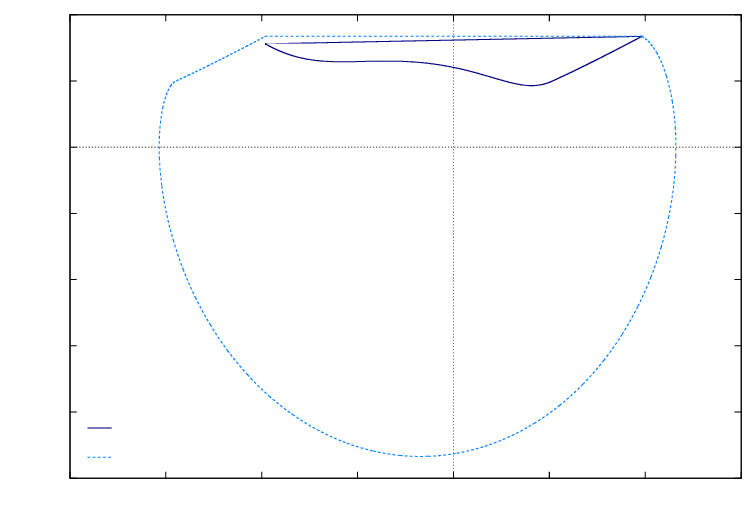}}
\vspace*{-1mm}
\caption{Phase-plane plot of steady walking motion}
\label{fig4.03}
\end{figure}

\begin{figure}[!t]
\hspace*{-9mm}
\scalebox{0.80}{
\input{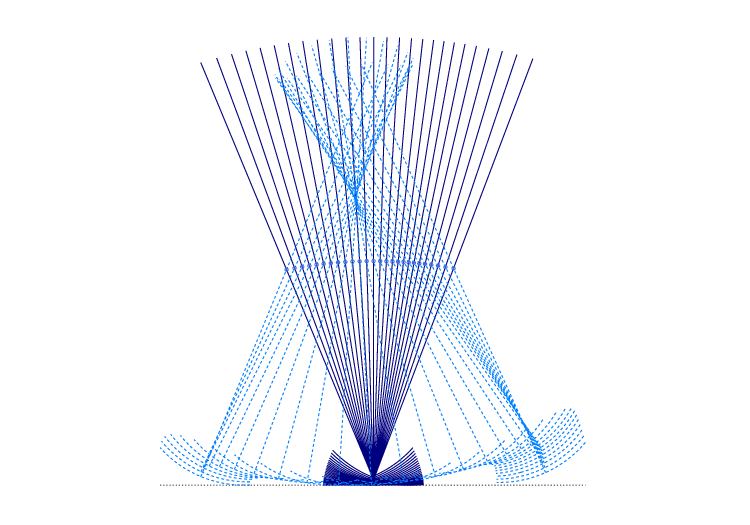}}\\
\vspace*{-9mm}
\caption{Stick diagram for one step of steady walking motion}
\label{fig4.04}
\end{figure}

\section{CONCLUSION AND FUTURE WORK}

In this paper, we proposed the planar X-shaped bipedal robot, and developed the mathematical model and output-following control system for stable level gait generation. The simulation results showed the validities of the proposed control method and special properties.

The semicircular-footed model has the feature that approximate linearization is effective when walking at low speed. In the future, we will further advance the mathematical analysis of gait descriptors such as $\Delta E_i$ and $\Delta T_\infty$ by using the linearized model, and build a method to know the gait stability and efficiency without relying on numerical simulation.

Now we are also investigating the superiority of low-speed limit cycle walking, especially its relationship to the magnitude of gravity. On the other hand, we are also investigating a method of generating another gait in which the swing leg does not overtake the stance leg or rotates in the same clockwise direction as the stance leg and swings down from above \cite{Kiefer,ICRA2010}. Such a gait is called a wheel gait, and it is expected that it will be effective in adapting to irregular terrain because it is not necessary to consider the clearance of the swing leg.

\end{document}